\documentclass[runningheads]{llncs}
\usepackage{ifxetex}
\usepackage{ntheorem}
\usepackage{amsmath}
\usepackage{amsfonts}
\ifxetex
    \usepackage{fontspec}
\else
    \usepackage[T1]{fontenc}
    \usepackage[utf8]{inputenc}
    \usepackage{lmodern}
\fi
\usepackage{graphics}
\usepackage{graphicx}
\usepackage{subcaption}
\usepackage{siunitx} 
\usepackage[misc]{ifsym}

\usepackage{tabularx} 
\newcolumntype{L}{>{$}l<{$}}                        % math-mode version of "l" column type (tabularx)
\newcolumntype{R}{>{$}r<{$}}                        % math-mode version of "r" column type (tabularx)
\newcolumntype{Y}{>{\centering\arraybackslash}X}    % centered X column (tabularx)
\newcolumntype{Z}{>{\raggedleft\arraybackslash}X}   % right-aligned X column (tabularx)

\usepackage{xcolor}

\usepackage{hyperref}
\hypersetup{
    bookmarksnumbered=true,bookmarksopen=true,
    unicode=true,colorlinks=true,linktoc=all,%linktoc=page
    linkcolor=blue,citecolor=blue,filecolor=blue,urlcolor=blue,
	pdfstartview=FitH
}
\usepackage{url}

\usepackage{algorithm}
\usepackage{algorithmic}

\usepackage{graphicx}
\usepackage{color}

\begin{document}
\title{Pattern Mining for Anomaly Detection in Graphs: Application to Fraud in Public Procurement}
\titlerunning{Pattern Mining in Graphs for Public Procurement Fraud Detection}
% \author{Lucas Potin \inst{1}\orcidID{0000-0003-1059-4095} \and
% Rosa Figueiredo\inst{1}\orcidID{0000-0002-0344-2686} \and
% Vincent Labatut\inst{1}\orcidID{0000-0002-2619-2835} \and
% Christine Largeron\inst{2}\orcidID{0000-0003-1059-4095}}
\author{Lucas Potin \inst{1} \and
Rosa Figueiredo\inst{1} \and
Vincent Labatut\inst{1} \and
Christine Largeron\inst{2}}

\authorrunning{L. Potin \textit{et al}.}

\institute{Laboratoire Informatique d'Avignon -- UPR 4128, F-84911, Avignon, France\\
\email{\{firstname\}-\{lastname\}@univ-avignon.fr} \and Laboratoire Hubert Curien -- UMR 5516, F-42023, Saint-Etienne, France
\email{christine.largeron@univ-st-etienne.fr}}
\maketitle              % typeset the header of the contribution
%
%%%%%%%%%%%%%%%%%%%%%%%%%%%%%%%%%%%%%%%%%%%%%%%%%%%%%%%%%%%%%%%%%
\begin{abstract} % 150--250 words
In the context of public procurement, several indicators called red flags are used to estimate fraud risk. They are computed according to certain contract attributes and are therefore dependent on the proper filling of the contract and award notices. However, these attributes are very often missing in practice, which prohibits red flags computation. Traditional fraud detection approaches focus on tabular data only, considering each contract separately, and are therefore very sensitive to this issue. In this work, we adopt a graph-based method allowing leveraging relations between contracts, to compensate for the missing attributes. We propose PANG (\underline{P}attern-Based \underline{An}omaly Detection in \underline{G}raphs), a general supervised framework relying on pattern extraction to detect anomalous graphs in a collection of attributed graphs. Notably, it is able to identify \textit{induced} subgraphs, a type of pattern widely overlooked in the literature. When benchmarked on standard datasets, its predictive performance is on par with state-of-the-art methods, with the additional advantage of being explainable. These experiments also reveal that induced patterns are more discriminative on certain datasets. When applying PANG to public procurement data, the prediction is superior to other methods, and it identifies subgraph patterns that are characteristic of fraud-prone situations, thereby making it possible to better understand fraudulent behavior.
\\[0.2cm]
\textcolor{red}{\textbf{Cite as:} L. Potin, R. Figueiredo, V. Labatut \& C. Largeron. ``Pattern Mining for Anomaly Detection in Graphs: Application to Fraud in Public Procurement'', \textit{European Conference on Machine Learning and Principles and Practice of Knowledge Discovery in Databases}, Springer, 2023. DOI: TBD}
\end{abstract}

%%%%%%%%%%%%%%%%%%%%%%%%%%%%%%%%%%%%%%%%%%%%%%%%%%%%%%%%%%%%%%%%%
\section{Introduction}
Public procurement refers to the purchase of goods, services and works by a public authority (the buyer), from a legal entity governed by public or private law (the winner). In the European Union, when the contract exceeds some price threshold, the buyer must first advertise a call for tenders defining its needs in detail, and later the corresponding award notice, which describes the content of the contract eventually concluded with one or more winners. These documents must be published in the \textit{Official Journal of the European Union} (OJEU). The online version of this journal, called the \textit{Tenders Electronic Daily} (TED)~\cite{TED2023}, publishes more than $650{,}000$ procurement notices a year. 
Consequently, the public procurement sector provides a huge amount of publicly available data.

Historically, anomalies in public procurement, which refer to doubtful behavior, are linked to specific characteristics associated with contracts. In the literature, these characteristics are called \textit{red flags}, and are used as indicators of potential fraud~\cite{Ferwerda2013,Rizzo2013,Fazekas2014,ferwerda2017}. For instance, modifying the contract price during the procedure, or receiving a single offer for a given call for tenders, are typically considered as red flags~\cite{OCP2016}. But the information required to compute these red flags is not always available. In the French subset of the TED, some essential attributes are largely missing~\cite{Potin2022}, e.g. the number of offers answering a call for tenders is not documented in 30\% of the cases. For such contracts, one can compute only \textit{partial} red flags, in the best of cases, or even no red flags at all.

Anomaly detection approaches are commonly used in fraud detection~\cite{Pourhabibi2020}. However, when applied to public procurement, most studies are based on \textit{tabular} data~\cite{Carvalho2013,Carneiro2020}, i.e. each contract is considered separately, as a set of attribute values. Only a very few authors try to take advantage of the \textit{relationships} between contracts by adopting a graph-based approach. Fazekas \& Tóth propose the CRI, a composite score combining several red flags, and leverage graphs~\cite{Fazekas2016}, but only to visualize its distribution over their dataset. Wachs \textit{et al}. \cite{Wachs2019} use graphs in order to estimate the proportion of red flags in the core agents, i.e. buyers and winners with the most frequent relationships, compared to the others. However, to the best of our knowledge, no method in the literature dedicated to anomaly or fraud detection in public procurement uses graphs to create predictive models. 

This leads us to propose a graph-based method to identify anomalies in public procurement. Our work makes three main contributions. First, we propose the \textit{PANG} framework (\underline{P}attern-Based \underline{An}omaly Detection in \underline{G}raphs), that leverages pattern mining to solve this problem. When evaluated on a benchmark of standard datasets, its performance is on par with state-of-the-art methods, with the additional advantage of being explainable. In addition, it allows looking for different types of patterns, including \textit{induced} subgraphs, which are generally overlooked in the literature. Our second contribution is to show empirically that such subgraphs can result in better classification performance on certain datasets. As a third contribution, we apply our generic framework to public procurement data, and identify the relevant patterns characterizing risky behaviors.

The rest of the article is structured as follows. Section~\ref{sec:RelatedWork} gives an overview of the literature regarding graph anomaly detection and graph pattern mining. Section~\ref{sec:ProblemFormulation} introduces the terminology used throughout this paper, as well as our problem formulation. Section~\ref{sec:RePPP} describes our framework PANG and assesses its performance on standard datasets. Section~\ref{sec:PublicProcurementUseCase} applies PANG to public procurement. Finally, we comment the main aspects of our work in Section~\ref{sec:Conclusion}. %, and identify its principal perspectives.

%%%%%%%%%%%%%%%%%%%%%%%%%%%%%%%%%%%%%%%%%%%%%%%%%%%%%%%%%%%%%%%%%
\section{Related Work}
\label{sec:RelatedWork}
The goal of anomaly detection is to detect behaviors significantly differing from expected norms. The methods dealing with this task on graphs either focus on single elements (vertices, edges) or larger structures (subgraphs, graphs) \cite{Akoglu2014,Ma2021,Kim2022}. When considering whole graphs, the task can be seen as a classification problem consisting in labelling the graph as normal or anomalous. The standard approach consists in building a vector-based representation of the graph, in order to apply classic data mining tools~\cite{Ma2021}. Most recent works focus on deep learning methods such as Graph Neural Networks (GNN)~\cite{Dou2021,Luo2022,Ma2022}, which not only learn this representation, but also tackle the classification task. However, one limitation of these methods lies in the lack of explainability: while some approaches have been proposed to make GNNs explainable~\cite{Yuan2022}, achieving this goal is non-trivial, especially when considering graphs with edge features. An alternative is to build the representation in a more controlled way, in order to retain its semantics~\cite{Yang2023}. Among the methods following this path, pattern-based approaches rely on the subgraphs that compose the graphs~\cite{AcostaMendoza2016}. They require retrieving the most characteristic of these patterns, generally the most frequent ones, in order to represent each graph in terms of absence or presence of these patterns.

There are different algorithms to extract \textit{frequent} subgraphs from a collection of graphs~\cite{Yan2002,FournierViger2019}, i.e. patterns appearing in more graphs than a fixed threshold. The main issue encountered with this approach is the pattern explosion problem, which states that the number of patterns increases exponentially when decreasing this threshold.
To alleviate the computational cost, some algorithms mine more constrained patterns, such as \textit{closed} frequent patterns~\cite{Shaul2021}, \textit{maximal} frequent patterns~\cite{Malik2023}, or \textit{approximate} patterns~\cite{Li2015}. As these notions are not the focus of this paper, we refer the reader to~\cite{Mooney2013} for further details. 

Moreover, all frequent patterns may not be relevant when dealing with a graph classification problem: some could occur equally in all classes, and thus provide no information to distinguish them. To overcome this issue, some methods have been proposed to mine \textit{discriminative} patterns. Leap~\cite{Yan2008} relies on a notion of structural proximity when building its search tree, that lets it compare branches in order to avoid exploring those that are similar. CORK~\cite{Thoma2010} is based on a metric that evaluates a pattern in relation to a collection of patterns already selected, which allows accounting for the proximity between frequent patterns. Moreover, this metric is submodular, and can thus be integrated into tools such as gSpan~\cite{Yan2002} to mine discriminative patterns efficiently. It also allows CORK to automatically select the number of patterns to extract. In~\cite{Kane2015}, the notion of discriminative pattern is extended in order to mine \textit{jumping emerging} patterns: subgraphs appearing in only one class. However, this notion is very restricted, as it requires that a pattern \textit{never} appears in one of the two classes. As a consequence, in practice, it often leads to very infrequent patterns~\cite{LoyolaGonzlez2020}. Our objective is to propose a generic classification framework which allows choosing the number of discriminative patterns to keep, as well as their type and, then to apply it for identifying fraud in public procurement.

%%%%%%%%%%%%%%%%%%%%%%%%%%%%%%%%%%%%%%%%%%%%%%%%%%%%%%%%%%%%%%%%%%
\section{Problem Formulation}
\label{sec:ProblemFormulation}
To detect fraud in public procurement, we adopt a network representation inspired by information retrieval or text mining, and previously successfully used for chemical compound classification~\cite{Metivier2015}. In the same way that a document can be modeled as a bag-of-words, we propose to represent a graph as a bag-of-subgraphs, i.e. the set of its constituting subgraphs, called \textit{patterns}. To do this, we construct a global dictionary constituted of the patterns appearing in a collection of attributed graphs. Based on this dictionary, each graph can then be represented as a fixed-length numerical vector, which can be used as an input by any standard machine learning algorithm.
In this section, we first describe how we define such vector-based representation, and then formulate our anomaly detection task as a classification problem.

\begin{figure}[hbt!]
    \centering
    \includegraphics[width=1\linewidth]{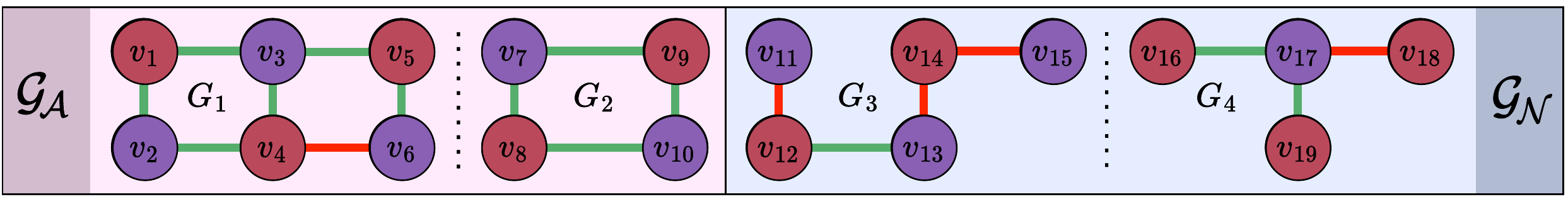}
    \caption{A collection $\mathcal{G}$ of graphs including the subsets of anomalous ($\mathcal{G}_A$) and normal ($\mathcal{G}_N$) graphs.}
    \label{fig:ExDataset}
\end{figure}

\begin{definition}[Attributed Graph]
    An attributed graph is defined as a tuple $G = (V,E,\mathbf{X},\mathbf{Y})$ in which $V$ is the set of $n$ vertices, $E$  the set of $m$ edges of $G$, $\mathbf{X}$ the $n \times d_v$ matrix whose row $\mathbf{x}_i$ is the $d_v$-dimensional attribute vector associated with vertex $v_i \in V$, and $\mathbf{Y}$ the $m \times d_e$ matrix whose row $\mathbf{y}_i$ is the $d_e$-dimensional attribute vector associated with edge $e_i \in E$.
\end{definition}
As an illustration, we consider a collection of such graphs, as shown in Figure~\ref{fig:ExDataset}. In this example, each vertex has an attribute corresponding to its color (brown or purple) as well as each edge (green or red).

Let us assume that each graph $G$ has a label $\ell_G$ picked in $\mathcal{L} =\{ A, N\}$, denoting an anomalous or a normal graph, respectively. Importantly, this label is not known for all the graphs at our disposal. Let $\mathcal{G}$ be the set of graphs whose label is known. The set $\mathcal{G}$ can be split into two disjoint subsets: $\mathcal{G} = \mathcal{G}_A \cup \mathcal{G}_N$ ($\mathcal{G}_A \cap \mathcal{G}_N = \emptyset$). Set $\mathcal{G}_A$ contains the anomalous graphs, and $\mathcal{G}_N$ the normal ones. Using the labeled set of graphs $\mathcal{G}$, our aim is to train a classifier able to predict the unknown label for the other graphs. For this purpose, we use a pattern-based graph representation.

\begin{figure}[hbt!]
    \begin{subfigure}[b]{0.31\textwidth}
        \centering
        \includegraphics[width=\textwidth]{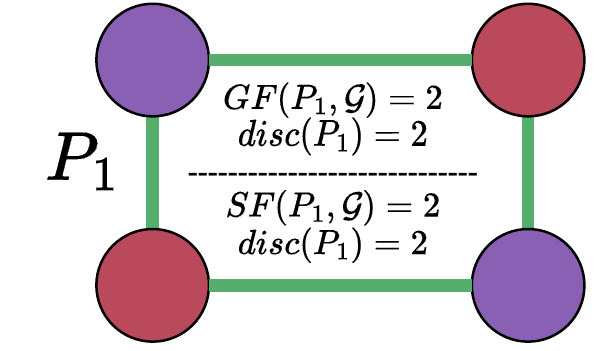}
    \end{subfigure}
    \hfill
    \begin{subfigure}[b]{0.31\textwidth}
        \centering
        \includegraphics[width=\textwidth]{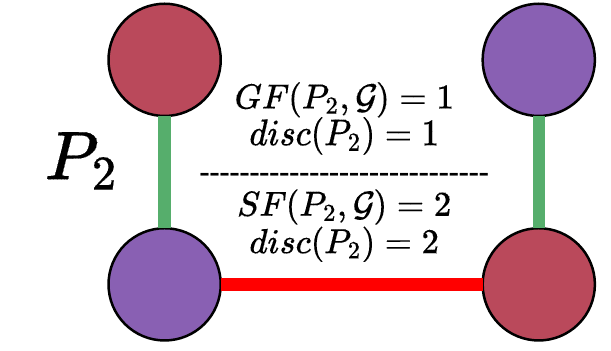}
    \end{subfigure} 
    \hfill
    \begin{subfigure}[b]{0.31\textwidth}
       \centering
        \includegraphics[width=\textwidth]{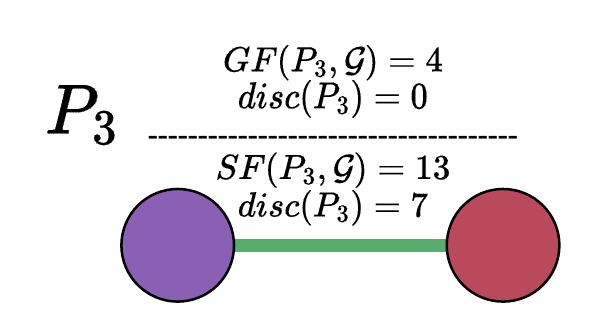}
    \end{subfigure} 
    \caption{Three examples of general patterns present in graph $G_1$ of Figure~\ref{fig:ExDataset}.}
    \label{fig:ExPattern}
\end{figure}

\begin{definition}[General Pattern]
    Let $G = (V,E,\mathbf{X},\mathbf{Y})$ be an attributed graph. A graph $P$ is a pattern of $G$ if it is isomorphic to a subgraph $H$ of $G$, i.e. $\exists H \subseteq G: P \cong H$.
    \label{def:Pattern}
\end{definition}
As we consider attributed graphs, we adopt the definition of a graph isomorphism proposed by Hsieh \textit{et al}.~\cite{Hsieh2006}, i.e. an isomorphism must preserve not only edges, but also vertex and edge attributes. We consider that $P$ is a pattern for a set of graphs $\mathcal{G}$ when $P$ is a pattern of at least one of its graphs. Figure~\ref{fig:ExPattern} shows three examples of patterns of $G_1$, and therefore of $\mathcal{G}$, from Figure~\ref{fig:ExDataset}.

It should be noted that, according to Definition~\ref{def:Pattern}, a pattern $P$ may not include all the edges originally present in $G$ between the considered vertices. We can restrict this definition by considering \textit{induced} patterns. Similarly to Definition~\ref{def:Pattern}, $P$ is an \textit{induced} pattern of $G$ if it is isomorphic to an \textit{induced} subgraph $H$ of $G$. %We use the definition of induced subgraph from Juttner \textit{et al}.~\cite{Juttner2018}.

\begin{definition}[Induced Subgraph]
    Let $G = (V, E, \mathbf{X}, \mathbf{Y})$ be an attributed graph. The subgraph $H = (V_H, E_H, \mathbf{X}_H, \mathbf{Y}_H)$ induced by a vertex subset $V_H \in V$ is such that $E_H = \{(u,v) \in E : u,v \in V_H\}$, and $\mathbf{X}_H$ and $\mathbf{Y}_H$ retain only the rows of $\mathbf{X}$ and $\mathbf{Y}$ matching $V_H$ and $E_H$, respectively.
    \label{def:Induced}
\end{definition}
In Figure~\ref{fig:ExPattern}, $P_1$ is an induced pattern of $G_1$. On the contrary, $P_2$ is a general pattern of $G_1$, but not an induced pattern, because edge $(v_3,v_5)$ from $G_1$ has no image in $P_2$. We consider that $P$ is an induced pattern of $\mathcal{G}$ when $P$ is an induced pattern of at least one of its graphs. To measure the importance of a pattern in $\mathcal{G}$, we now need the notion of \textit{graph frequency}.

\begin{definition}[Graph Frequency]
    The graph frequency $GF(P,\mathcal{G})$ of a pattern $P$ in $\mathcal{G}$ is the number of graphs in $\mathcal{G}$ having $P$ as a pattern: 
    
    $GF(P,\mathcal{G}) = | \{ G \in \mathcal{G} : \exists H \subseteq G \text{ s.t. } P \cong H \} |$.
\end{definition}
It indicates the number of graphs having a specific pattern, but does not give any information about the number of times the pattern appears in these graphs. For this, we use the \textit{subgraph frequency}.
\begin{definition}[Subgraph Frequency]
    The subgraph frequency $SF(P,\mathcal{G})$ of a pattern $P$ in $\mathcal{G}$ is its total number of occurrences over all $G \in \mathcal{G}$: 
    
    $SF(P,\mathcal{G}) = \sum_{G \in \mathcal{G}} | \{ H \subseteq G : P \cong H \} |$.
\end{definition}

Graph frequency can be used to define the notion of \textit{closed} pattern, which in turn allows finding a more compact set of relevant patterns.

\begin{definition}[Closed Pattern]
    A pattern $P$ of $\mathcal{G}$ is said to be \textit{closed} if it has no supergraph $P'$, or equivalently if $P$ is not the subgraph of any graph $P'$, such that $GF(P',\mathcal{G}) = GF(P,\mathcal{G})$.
\end{definition}
As a consequence, the set of closed patterns is a subset of the set of general patterns. In our example, there is no supergraph of $P_1$ appearing in $2$ graphs, which makes it a closed pattern of $\mathcal{G}$. 

Regardless of the type of pattern, we note $\mathcal{P}_A$ and $\mathcal{P}_N$ the sets of patterns of $\mathcal{G}_A$ and $\mathcal{G}_N$, respectively, and $\mathcal{P}$ the complete set of patterns of $\mathcal{G}$: $\mathcal{P} = \mathcal{P}_A \cup \mathcal{P}_N$. Not all patterns are equally relevant to solve a given task. For instance, in Figure~\ref{fig:ExPattern}, $P_3$ is much more common than both other patterns in $\mathcal{G}$ from Figure~\ref{fig:ExDataset}. To distinguish them, we rely on the discrimination score from~\cite{Thoma2010}, that characterizes each pattern according to its frequency in the two subsets.

\begin{definition}[Discrimination Score]
    The discrimination score of a pattern $P$ of $\mathcal{G}$ is defined as $disc(P) = \displaystyle |F(P,\mathcal{G}_A) - F(P,\mathcal{G}_N)|$, where $F$ is $GF$ or $SF$.
    \label{def:Score}
\end{definition}
Our definition generalizes that of~\cite{Thoma2010}, so that it can be applied to both frequencies ($GF$ and $SF$). A score close to 0 indicates a pattern that is as frequent in $\mathcal{G}_A$ as in $\mathcal{G}_N$, while a higher score means that the pattern is more frequent in one of the two subsets. 
We use this score to rank the patterns in $\mathcal{P}$, and select the $s$ most discriminative ones ($1 \leq s \leq |\mathcal{P}|$). Some methods, like CORK~\cite{Thoma2010}, estimate $s$ automatically, which can be an advantage or a drawback, depending on the level of control desired by the user.

The resulting subset $\mathcal{P}_s \subseteq \mathcal{P}$ constitutes our dictionary, which means that $s$ lets us control the dimension of our graph representation. The representation of each graph $G_i \in \mathcal{G}$ is a vector $\mathbf{h}_i \in \mathbb{R}^{s}$ whose components measure how important each pattern of $\mathcal{P}_s$ is to $G_i$. These measures can be computed according to different formula, as discussed in Section~\ref{sec:RePPP}. Finally, we build the matrix $\mathbf{H} \in \mathbb{R}^{|\mathcal{G}| \times s}$ by considering the vector representations of all the graphs in $\mathcal{G}$. 

Based on this graph representation, our anomaly detection problem amounts to classifying graphs with unknown labels as anomalous or normal. More formally, given the training set composed of a set of graphs $ \mathcal{G} = \{G_i, i = 1, \dots, |\mathcal{G}|\}$ with the labels $\ell_{G_i} \in \mathcal{L}$ and the vector representations $\mathbf{h_i}$, the goal is to  learn a function $f : \mathbb{R}^{s} \rightarrow \{A,N\}$, which associates a label (anomalous or normal) to the vector representation of an unlabeled graph.

%%%%%%%%%%%%%%%%%%%%%%%%%%%%%%%%%%%%%%%%%%%%%%%%%%%%%%%%%%%%%%%%%%
\section{PANG Framework}
\label{sec:RePPP}

%%%%%%%%%%%%%%%%%%%%%%%%%%%%%%%%%%
\subsection{Description of the Framework}
\begin{figure}[hbt!]
    \centering \includegraphics[width=\textwidth]{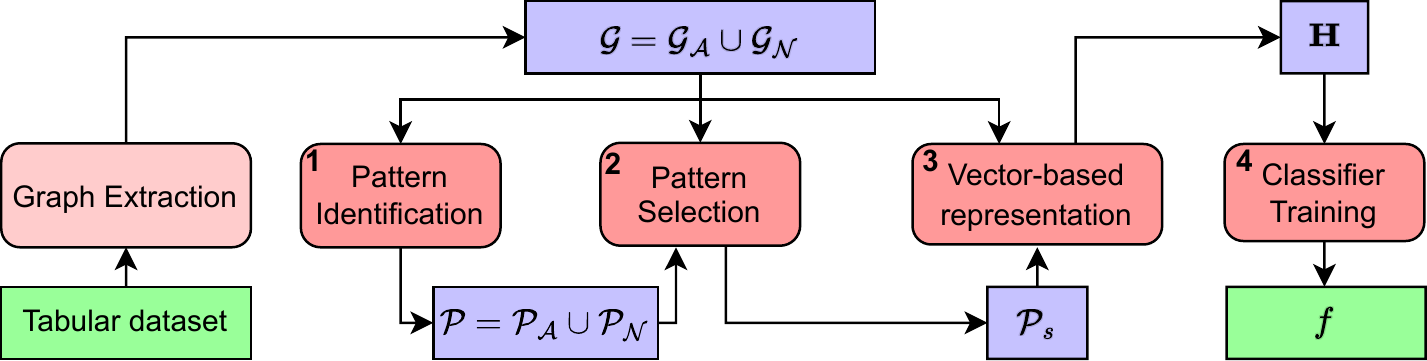}
    \caption{Processing steps of the proposed PANG framework.}
    \label{fig:Flowchart}
\end{figure}

To solve our classification problem, we propose the PANG framework (\underline{P}attern-Based \underline{An}omaly Detection in \underline{G}raphs), whose source code is publicly available online\footnote{\url{https://github.com/CompNet/Pang/releases/tag/v1.0.0} \label{ftn:pang}}. A preliminary step consists in extracting the graphs, but as it is data-dependent, we defer its description to Section~\ref{sec:graphConstruction}. The rest of the process is constituted of four steps, as represented in Figure~\ref{fig:Flowchart}:
\begin{enumerate}
    \item Identify all the patterns of $\mathcal{G}$ and build $\mathcal{P}$.
    \item Select the most discriminative patterns $\mathcal{P}_s$ among them.
    \item Use these patterns to build the vector-based representation of each graph.
    \item Train a classifier to predict the graph labels based on these representations.
\end{enumerate}

%%%
\paragraph{Step \#1: Pattern Identification}
\label{sec:FP}
In order to create $\mathcal{P}$, we use an existing graph pattern extractor. Several tools are available to enumerate patterns, such as gSpan~\cite{Yan2002}, FFSM~\cite{Huan2003}, or more recently TKG~\cite{FournierViger2019} and cgSpan~\cite{Shaul2021}. 

gSpan and cgSpan respectively search the \textit{frequent} and \textit{closed frequent} patterns in a set of graphs. Both rely on an iterative procedure, which starts from the simplest pattern possible, i.e. a single vertex with a specific attribute, in order to initialize the list of ranked frequent patterns. At each step, the algorithm takes the most frequent pattern according to this list, and tries to extend it by adding an edge. This expansion results in a set of new patterns, which are added or not to the ranked list, according to their frequency. This list is updated over the iterations, until it is no longer possible to find any new pattern with a frequency potentially higher than a predefined threshold.

In the case of cgSpan, the algorithm is able to find the set of closed frequent patterns, which, as explained before, is included in the set of frequent patterns. A smaller set of patterns allows reducing the computation time during the pattern mining phase, but also at post-processing, e.g. when computing the discrimination scores, since there are fewer patterns to consider, and consequently a smaller size for the vector representation.

We choose to use gSpan~\cite{Yan2002} and cgSpan~\cite{Shaul2021}. The former mines an important number of frequent patterns while requiring less memory than TKG. The latter is able to efficiently identify closed patterns. Both algorithms are implemented in Java, and are available as a part of software SPMF~\cite{FournierViger2016}, which provides numerous tools for pattern mining.
The process used for the induced patterns is based on two steps: first, each pattern is extracted using one of these algorithms. Then, we filter the induced patterns using the ISMAGS algorithm~\cite{Houbraken2014} implemented in NetworkX~\cite{Hagberg2008}.

%%%
\paragraph{Step \#2: Discriminative Pattern Selection}
Next, we compute the discrimination score of each extracted pattern as explained in Definition~\ref{def:Score}. We keep the $s$ most discriminative patterns to construct $\mathcal{P}_s$. 
%The purpose of this parameter is to limit the size of the representation space, and therefore the number of patterns collected.

%%%
\paragraph{Step \# 3: Vector-Based Representation}
\label{sec:PBRep}
Once we have $\mathcal{P}_s$, we compute the vector representation of each graph in $\mathcal{G}$. In this work, we use several approaches. First, we build a binary vector indicating the presence or absence of each pattern in the considered graph. In that case, for each graph $G_i \in \mathcal{G}$ and each pattern $P_j \in \mathcal{P}$, $H_{ij}$ equals $1$ if this pattern $P_j$ is present in $G_i$ and $0$ otherwise.

This representation is somewhat limited, though, as it ignores how much patterns are present in graphs. To solve this issue, we propose an integer representation based on the number of occurrences \textit{in the graph}. This number is computed with the VF2 algorithm~\cite{Cordella2004}, available in Networkx~\cite{Hagberg2008}. Given a pattern $P$ and a graph $G$, VF2 identifies the number of subgraph isomorphisms of $P$ in $G$, which we store in $H_{ij}$.

\begin{figure}[hbt!]
    \centering
    \includegraphics[width=1\textwidth]{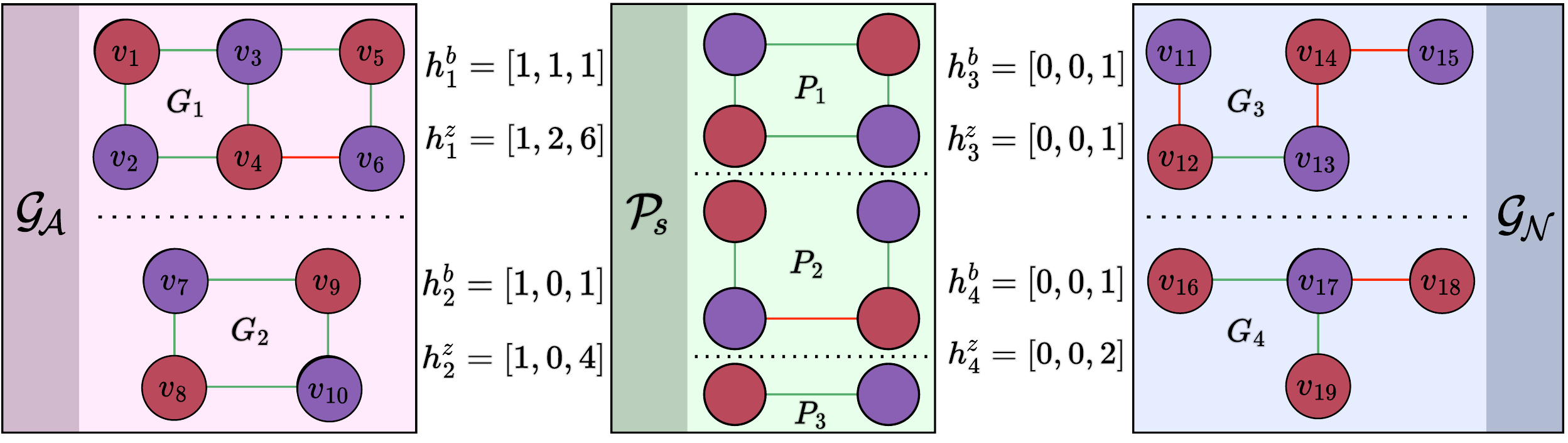}
    \caption{Binary ($\mathbf{h}_j^b$) and integer ($\mathbf{h}_j^z$) vector-based representations of the graphs of Figure~\ref{fig:ExDataset}, using the patterns of Figure~\ref{fig:ExPattern} as $\mathcal{P}_s$.}
    \label{fig:vector}
\end{figure}

Figure~\ref{fig:vector} shows the representations obtained for the graphs of Figure~\ref{fig:ExDataset}, using the patterns from Figure~\ref{fig:ExPattern} as $\mathcal{P}_s$. Vectors $h^{b}_j$ and $h^{z}_j$ denote the binary and integer representations of each graph $G_j$, respectively.
It is worth noting that two different graphs can have the same vector representation, as is the case for the binary representation of $G_3$ and $G_4$ in our example.

For the sake of consistency, we compute the discrimination scores based on $GF$ when using the binary representation, and on $SF$ when using the integer one.

%%%
\paragraph{Step \#4: Classifier Training}
\label{sec:ClassifierTraining}
After the previous step, each graph is represented by a fixed-sized vector, no matter its number of vertices or edges. We leverage this representation to train a classifier into predicting the graph labels. Our framework is general and allows any classifier, but we select C-SVM~\cite{Chang2011} in this article, as it gives the best experimental results (cf. Appendix~\ref{app:classifier} for more classifiers).

%%%%%%%%%%%%%%%%%%%%%%%%%%%%%%%%
\subsection{Assessment on Benchmarks}
\label{sec:Benchmark}
Before focusing on fraud detection in public procurement, we assess PANG on FOPPA, the public procurement dataset that we use in our application, as well as four real-world datasets commonly used in the literature as benchmarks. These last datasets, our protocol and the results of this first experiment are detailed in the following. The FOPPA data is described in Section~\ref{sec:graphConstruction}.

%%%
\paragraph{Experimental Protocol}
MUTAG~\cite{Debnath1991} contains $188$ graphs representing molecules, where vertices are atoms and edges bonds between them. The graphs are distributed over two classes, depending on the molecule mutagenicity. PTC\_FR~\cite{Toivonen2003} contains $350$ graphs, also representing molecules. There are also two graph classes, depending on the molecule carcinogenicity on male and female mice and rats. NCI1~\cite{Wale2006} contains $4{,}110$ graphs representing chemical compounds. Each vertex stands for an atom, while edges represent the bonds connecting them. Like before, there are two classes distinguishing the compounds depending on their carcinogenicity. D\&D~\cite{Dobson2003} is composed of $1{,}178$ protein structures. Each vertex is an amino acid, and two vertices are connected if they are less than 6 angstroms apart. There are two graph classes corresponding to enzymes vs. non-enzymes. Table~\ref{tab:BenchMarkDataset} shows the main characteristics of these datasets: number of graphs, and average numbers of vertices and edges.

\begin{table}[hbt!]
    \begin{center}
        \tabcolsep = 2\tabcolsep
        \begin{tabularx}{\textwidth}{X r r r r}
            \hline
            \textbf{Datasets} & \textbf{MUTAG} & \textbf{PTC\_FR} & \textbf{NCI1} & \textbf{D\&D} \\
            \hline
            Number of graphs & $188$ & $350$ & $4{,}110$ & $1{,}178$ \\
            Average number of vertices & $17.93$ & $25.56$ & $29.87$ & $284.72$ \\
            Average number of edges & $19.79$ & 25.96 & $32.30$ & $715.66$ \\
            \hline
        \end{tabularx}
        \caption{Characteristics of the 4 benchmark datasets.} \label{tab:BenchMarkDataset}
    \end{center}
\end{table}

Regarding graph representations, we compute the six types proposed in PANG:
\begin{itemize}
    \item PANG\_GenBin: binary representation considering general patterns.
    \item PANG\_GenOcc: integer representation considering general patterns.
    \item PANG\_IndBin: binary representation using only induced patterns.
    \item PANG\_IndOcc: integer representation using only induced patterns.
    \item PANG\_CloBin: binary representation using only closed patterns.
    \item PANG\_CloOcc: integer representation using only closed patterns. 
\end{itemize}

% 1) PANG
% 2) Kernels
% 3) Graph2VEC 
% ----> recuperer la representation et classifier après
%4) DGCNN : classifie directement.
We compare our results with four different types of baselines. First, as an alternative pattern-based method, we use CORK (cf. Section~\ref{sec:RelatedWork}), which automatically estimates the size of the representation. The second baseline type is graph kernels. We use the kernel matrices of the graphs as representations, associating each row of the matrix with the corresponding graph. These matrices are computed from the implementation of the WL kernel~\cite{Shervashidze2011} and the WL\_OA~\cite{Kriege2016} kernel, both available in the GraKel~\cite{Siglidis2020} library. The third type is whole graph embedding neural methods, for which we use Graph2Vec~\cite{Annamalai2017}, available in the KarateClub library~\cite{Rozemberczki2020}. We set an embedding size of $128$, which is standard in the literature. For each of these representations, we train a C-SVM as indicated in Step 4 of Section~\ref{sec:ClassifierTraining}.

The fourth baseline type is Graph Neural Networks, with DGCNN~\cite{Zhang2018}. This method produces a graph representation, which can be fetched to the SVM, but it can also perform the classification step directly. The results reported here are the best ones, obtained in this second setting, using the implementation from StellarGraph~\cite{CSIRO2018}, with the optimal parameter values as indicated in~\cite{Zhang2018}.

%%%
\paragraph{Experimental Results}
We adopt a 10-fold cross-validation to assess classifier performance. Table~\ref{tab:BenchmarkRes} shows the average $F$-Score (with standard deviation) for the \textit{Anomalous} class. Each column corresponds to one of the considered datasets: 4 benchmarks and FOPPA.

\begin{table}[hbt!]
    \begin{center}
        \tabcolsep = 2\tabcolsep
        \begin{tabularx}{\textwidth}{X r r r r r}
            \hline
            Representation & \textbf{MUTAG} & \textbf{NCI1} & \textbf{D\&D}  & \textbf{PTC} & \textbf{FOPPA} \\
            \hline
            PANG\_GenBin  & 0.85 (0.05) & 0.79 (0.02) & 0.77 (0.03) & 0.60 (0.13) & 0.93 (0.02) \\ 
            PANG\_GenOcc  & 0.87 (0.04) & 0.77 (0.02) & 0.75 (0.02) & 0.58 (0.10) & 0.91 (0.03) \\ 
            PANG\_IndBin  & 0.87 (0.05) & 0.79 (0.02) & 0.76 (0.03) & 0.59 (0.13) & \textbf{0.95 (0.01)} \\ 
            PANG\_IndOcc  & 0.87 (0.03) & 0.79 (0.01) & 0.75 (0.03) & 0.56 (0.07) & 0.92 (0.02)  \\ 
            PANG\_CloBin & 0.86 (0.05) & 0.78 (0.03) & 0.75 (0.03)& 0.57 (0.15) & 0.94 (0.03)\\
            PANG\_CloOcc & \textbf{0.88 (0.04)} & 0.76 (0.02) & 0.71 (0.04) & 0.54 (0.11) & 0.92 (0.02)\\
            \hline
            CORK  & 0.66 (0.08) & 0.78 (0.02) & 0.73 (0.03) & 0.54 (0.06) & 0.63 (0.05)  \\ 
            WL & 0.86 (0.06) & \textbf{0.83 (0.01)} & \textbf{0.82 (0.01)} & 0.57 (0.06) & 0.90 (0.05) \\
            WL\_OA  & 0.86 (0.06) & 0.81 (0.03) & 0.77 (0.03) & 0.55 (0.11) & 0.90 (0.05) \\ 
            Graph2Vec & 0.84 (0.07) & 0.82 (0.01) & 0.72 (0.03) & \textbf{0.61 (0.11)} & 0.91 (0.04) \\
            DGCNN & 0.86 (0.04) & 0.74 (0.01) & 0.79 (0.01) & 0.58 (0.05) & 0.89 (0.01) \\
            \hline
        \end{tabularx}
        \caption{$F$-Scores ($\pm$ standard deviation) for the \textit{Anomalous} class.} 
        \label{tab:BenchmarkRes}
    \end{center}
\end{table}

No method dominates the others over all datasets, therefore we can assume that some graph representations are more relevant to model certain systems. We plan to investigate this question further, but this is out of this article's scope. 
The performance of PANG is systematically above CORK, its most similar method. This is because, on the considered datasets, CORK identifies a very restricted set of discriminative patterns and trades classification performance against representation size.
Moreover, PANG is on par with the remaining methods on NCI1, D\&D and PTC, and has the best performance on MUTAG and, importantly, on FOPPA, our application dataset. Thus, we assume that PANG is able to capture the same information as embedding- and GNN-based methods. On the one hand, it requires numerous patterns to be mined, and is therefore more time-consuming than these methods. On the other hand, it has the advantage of being interpretable, allowing us to identify the most discriminative patterns. This is why we apply it to fraud detection in public procurement, in Section~\ref{sec:PublicProcurementUseCase}.

%%%%%%%%%%%%%%%%%%%%%%%%%%%%%%%%%%%%%%%%%%%%%%%%%%%%%%%%%%%%%%%%%%
\section{Public Procurement Use Case}
\label{sec:PublicProcurementUseCase}
In this section, we apply PANG to real data representing public procurement. We first describe the process used to extract graphs from a database of French public procurement contracts (Section~\ref{sec:graphConstruction}), then we discuss our results (Section~\ref{sec:ProcResults}).

%%%%%%%%%%%%%%%%%%%%%%%%%%%%%%
\subsection{Extraction of the Graph Dataset}
\label{sec:graphConstruction}

%%%
\paragraph{Raw Data}
The FOPPA~\cite{Potin2022} database lists all French contracts award notices published at the European level. Each such contract involves \textit{at least} two economic \textit{agents}: a buyer and a winner, and may be constituted of several lots. It is described by a collection of attributes such as the total price, the number of offers, the bid ranking criteria, and whether the procedure was accelerated. %Note that it is possible for an entity to be a buyer and a winner in different contracts. 
In this paper, we consider the specific subset of contracts concerning period 2015--19, containing $417{,}809$ lots.

%%%
\paragraph{Contract Filtering}
We could apply our graph extraction process to the whole set of French contracts, however this would result in a single graph, combining heterogeneous activity domains and agent types. Yet, some attributes, for example the weight of social and environmental criteria, directly depend on these domains and types~\cite{Marchal2022}. Instead, we select only a part of the available data to constitute a collection of consistent contracts. For this purpose, we filter them according to five aspects: agent category, activity sector, temporal period, geographic region and size. Regarding the agents, we focus on municipalities, because they are very numerous, and automating their identification is more straightforward than for the other types of public agents. For each municipality present in the dataset, we build a subset of contracts containing not only its own contracts, but also those involving their winners, as well as the other municipalities with which they have obtained contracts. The other four filters allow us to control the size of these subsets of contracts, while retaining a certain homogeneity: we keep only those related to works, covering periods of one year, and involving only suppliers belonging to the same French administrative subdivision.

After this filtering, we obtain a collection of contract subsets containing a total of $25{,}252$ contracts. For each contract, we compute a standard red flag from the literature, in order to model how fraudulent it could be. A contract is red flagged if the number of offers received is exactly 1, which reveals a lack of competition~\cite{OCP2016}.

%%%
\paragraph{Graph Extraction}
For each contract subset obtained after the filtering, we extract a graph $G$. We consequently build a set of graphs, corresponding to $\mathcal{G}$ in Section~\ref{sec:RePPP}. In the context of public procurement, due to the complexity of the data, one can extract various types of graphs~\cite{Fazekas2016}, depending on what the vertices, edges, and their attributes, represent.

We use vertices to model agents, and edges to represent relationships between them, i.e. their joint involvement in at least one contract. Each vertex has an attribute, indicating whether the agent is a buyer or a winner, while each edge has an attribute related to the number of lots contracted between a buyer and a winner. We limit the latter to three levels: 1) exactly one lot; 2) between $2$ and $5$ lots; and 3) $6$ lots or more. This allows us to identify cases where a buyer has many contracts with a single winner, a behavior generally associated with red flags in the literature~\cite{FalcnCorts2022}.

We consider that an edge is anomalous if it represents \textit{at least} one red flagged contract, i.e. a contract that received exactly one offer. The label of a graph depends on its total number of anomalous edges: normal if there are fewer than $2$, anomalous otherwise.
%
%
%%%
%\paragraph{FOPPA Dataset Properties}
Our graph extraction method produces $389$ normal and $330$ anomalous graphs. 
Table~\ref{tab:Dataset} shows the main characteristics of the resulting FOPPA dataset, which is publicly available online with our source code\footnote{\url{https://github.com/CompNet/Pang/releases/tag/v1.0.0}}.

\begin{table}[hbt!]
    \begin{center}
        \tabcolsep = 2\tabcolsep
        \begin{tabularx}{\textwidth}{X r r}
            \hline
            \textbf{Graph} & \textbf{Average number} & \textbf{Average number} \\
            \textbf{Class} & \textbf{of vertices (std)} & \textbf{of edges (std)} \\
            \hline
            \textit{Anomalous} & 15.76 (5.56) & 17.09 (7.86) \\
            \textit{Normal}    & 12.54 (5.41) & 12.59 (6.90) \\
            \hline
        \end{tabularx}
        \caption{Characteristics of the graphs extracted from the FOPPA dataset.} 
        \label{tab:Dataset}
    \end{center}
\end{table}

%%%%%%%%%%%%%%%%%%%%%%%%%%%%%%%%%%%%
\subsection{Results on Public Procurement Data}
\label{sec:ProcResults}

%%%
\paragraph{Comparison With a Tabular Representation} 
In order to study the impact of our graph-based representations, we compare them to a baseline relying on the traditional \textit{tabular} approach. For each contract, we use as predictive features $15$ fields available in FOPPA. We select only relevant fields such as the type of procedure, or the presence of a framework agreement. With these features, we aim to predict a binary class, based on the same red flag as before: the number of offers for the contract. Class $0$ contains the contracts with more than $1$ tender, and Class $1$ those with a unique tender. Note that the predictive features are independent from the number of offers.

Like for the graphs, we train an SVM with 10-fold cross-validation, on the same $25{,}252$ contracts obtained after the filtering step. However, the resulting prediction is defined at the \textit{contract} level (one row in the tabular data), whereas PANG works at the agent level (one graph in the collection). To compare these results, we need to group the tabular predictions by agent. For this purpose, we proceed as in Section~\ref{sec:graphConstruction}, by considering any agent with two red flagged contracts or more as anomalous.

\begin{table}[hbt!]
    \begin{center}
        \tabcolsep = 2\tabcolsep
        \begin{tabularx}{\textwidth}{X r r}
            \hline
            \textbf{Type of data} & \textbf{\textit{Anomalous} Class} & \textbf{\textit{Normal} Class} \\
            \hline
            Tabular Data & 0.19 (0.01) & 0.66 (0.01) \\
            PANG\_IndBin & \textbf{0.95 (0.01)} & \textbf{0.93 (0.02)} \\ 
            \hline
        \end{tabularx}
        \caption{$F$-Scores ($\pm$ standard deviation) for both classes, obtained with the tabular and graph data.} 
        \label{tab:ResTabular}
    \end{center}
\end{table}

Table~\ref{tab:ResTabular} compares the obtained performance with our best graph-based results. The $F$-Scores are averaged over the 10 folds, with standard deviation, for the \textit{Anomalous} and \textit{Normal} classes.
For the same contracts and classifier (C-SVM), the graphs allow us to predict fraudulent behaviors much more efficiently than the tabular data, notably for anomalous agents. This clearly confirms the interest of taking advantage of relationships between agents to tackle fraud detection, especially when red flags are missing. %This also raises the question whether combining both types of information would improve the performance even more, which we plan to explore in the future.

\begin{figure}[hbt!]
\centering
\includegraphics[width=\textwidth]{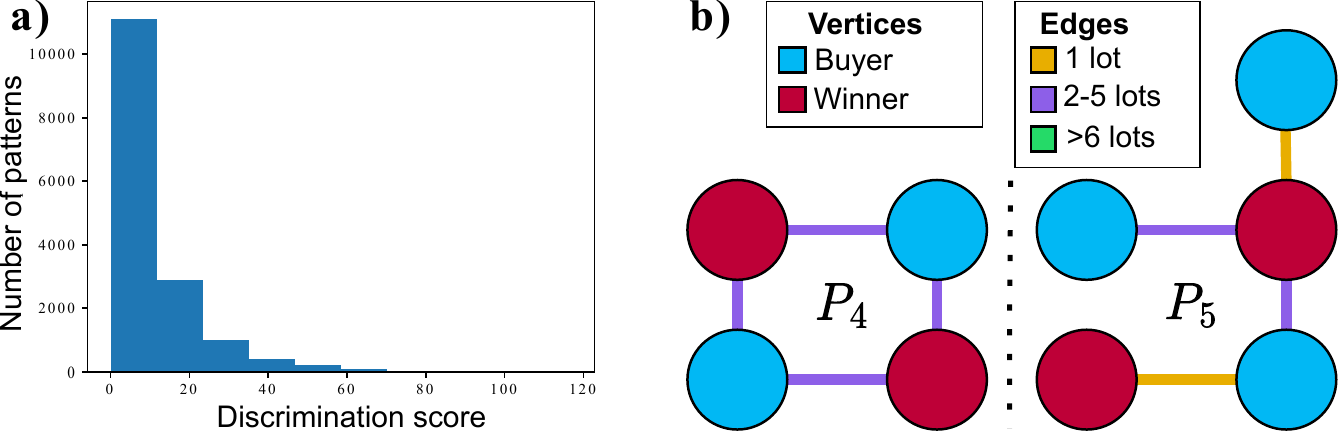}
    \caption{(a) Distribution of the patterns in function of their discrimination scores
 (b) Examples of discriminative patterns.}
    \label{fig:Resultats}
\end{figure}

%%%
\paragraph{Discrimination Score}
When applied to our dataset, gSpan returns a total of $15{,}793$ distinct patterns. Figure~\ref{fig:Resultats}.a shows the distribution of their discrimination score. It is in $[0;20]$ for most patterns ($85\%$), which can thus be considered as non-discriminative.
Figure~\ref{fig:Resultats}.b shows examples of 2 discriminative patterns, with respective scores of $64$ and $91$. Both of them include several relations with an intermediary number of lots, which are rather common in large graphs, and more often associated with anomalous graphs.

%%%
\paragraph{Impact of the Number of Discriminative Patterns}
We now study how the performance is affected by the number $s$ of patterns in $\mathcal{P}_s$, i.e. the vector representation size. 
Table~\ref{tab:PatternSize} shows how the $F$-Score changes depending on $s$, for anomalous and normal graphs. The last row indicates the performance obtained with all the identified patterns ($s = |\mathcal{P}|$). A representation based on only $100$ patterns, i.e. less than $1\%$ of the  $15{,}793$ patterns, is sufficient to reach the $0.8$ bar for both classes. This represents around $90\%$ of the maximal $F$-Score, obtained with all patterns. Therefore, only a small number of patterns are required to convey the information necessary to tackle the classification task. % This result can be leveraged to speed up the pattern mining step.

\begin{table}[hbt!]
    \begin{center}
        \tabcolsep = 2\tabcolsep
        \begin{tabularx}{\textwidth}{Z Z Z}
            \hline
            % \textbf{Number of} & \textbf{\textit{Anomalous} Class} & \textbf{\textit{Normal} Class} \\
            % \textbf{patterns in $\mathcal{P}_s$} & \textbf{$F$-Score}&\textbf{$F$-Score} \\
            \textbf{Representation Size} $s$ & \textbf{\textit{Anomalous} Class} & \textbf{\textit{Normal} Class} \\
            \hline
            10 & 0.66 (0.05) & 0.73 (0.05)\\
            50 & 0.74 (0.05) & 0.77 (0.04)\\
            100 & 0.81 (0.05) & 0.83 (0.04) \\
            150 & 0.88 (0.03) & 0.88 (0.03) \\
            (\textit{all}) 15,793 & \textbf{0.93 (0.02)} & \textbf{0.93 (0.02)}\\
       \hline
        \end{tabularx}
        \caption{$F$-Score ($\pm$ standard deviation) depending on parameter $s$, the size of $\mathcal{P}_s$.} 
        \label{tab:PatternSize}
    \end{center}
\end{table}

%%%
\paragraph{Impact of the Type of Patterns}
We also study how the type of pattern influences the constitution of $\mathcal{P}_s$, and therefore the classification performance. For this purpose, we set $s = 100$, and compare the six representations proposed by PANG, as we did in Section~\ref{sec:Benchmark}.
Table~\ref{tab:PatternType} shows the $F$-Score obtained with each representation, for both classes.

\begin{table}[hbt!]
    \begin{center}
        \tabcolsep = 2\tabcolsep
        \begin{tabularx}{\textwidth}{X r r}
            \hline
            % \textbf{Type of} & \textbf{\textit{Anomalous} Class} & \textbf{\textit{Normal} Class} \\
            % \textbf{representation} & \textbf{$F$-Score}&\textbf{$F$-Score} \\
            \textbf{Representation Type} & \textbf{\textit{Anomalous} Class} & \textbf{\textit{Normal} Class} \\
            \hline
            PANG\_GenBin & 0.81 (0.05) & 0.83 (0.04) \\
            PANG\_GenOcc & 0.73 (0.07) & 0.79 (0.05) \\
            PANG\_IndBin & \textbf{0.84 (0.03)} & \textbf{0.85 (0.03)} \\
            PANG\_IndOcc & 0.82 (0.05) & 0.84 (0.04) \\
            PANG\_CloBin & \textbf{0.84 (0.04)} & \textbf{0.85 (0.04)} \\
            PANG\_CloOcc & 0.83 (0.05) & \textbf{0.85 (0.04)}\\
            \hline
        \end{tabularx}
        \caption{$F$-Score ($\pm$ standard deviation) depending on the pattern type of the representation.}
        \label{tab:PatternType}
    \end{center}
\end{table}

Representations based on induced and closed patterns systematically lead to better results. Yet, a manual examination of $\mathcal{P}_s$ reveals that the discrimination scores of their selected patterns are similar to the general case. The worst selected pattern reaches a score of $67$ for general patterns, vs. $61$ for induced and $64$ for closed patterns. The difference lies in the \textit{nature} of the selected patterns, which are more diverse than when mining general patterns. For induced and closed patterns, $\mathcal{P}_s$ includes respectively $16$ and $13$ patterns that do not appear when dealing with general patterns.

\begin{figure}[hbt!]
    \centering         
    \includegraphics[width=1\textwidth]{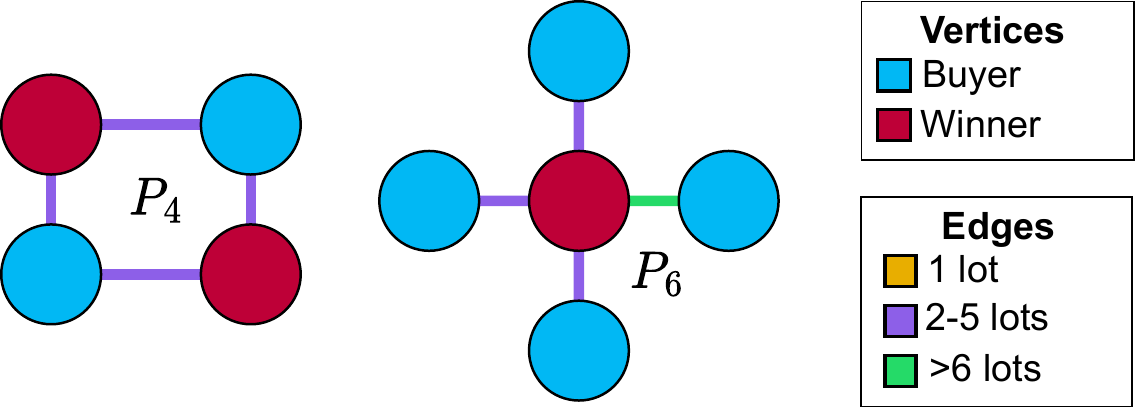} 
    \caption{Examples of discriminative patterns characteristic of class \textit{Anomalous}.}
    \label{fig:privilege}
\end{figure}

%%%
\paragraph{Interpretation of Fraudulent Behavior through Pattern Analysis}
An important advantage of our framework is the identification of the most discriminative patterns, and thus the possibility to leverage human expertise to interpret these patterns and better understand the reasons why an agent is considered fraudulent. For illustration, Figure~\ref{fig:privilege} shows two discriminative patterns, $P_4$ and $P_6$. %, according to our own score. 
Pattern $P_4$ represents a relationship between two winners and two buyers, with more than one contract between them. This type of pattern occurs more frequently in graphs with more contracts, which is typical of anomalous graphs. Pattern $P_6$ has a winner connected to several buyers, and a single of these edges is green. This can be interpreted as favoritism: a winner works much more with a municipality than with the others.

%%%%%%%%%%%%%%%%%%%%%%%%%%%%%%%%%%%%%%%%%%%%%%%%%%%%%%%%%%%%%%%%%%
\section{Conclusion}
\label{sec:Conclusion}
In this paper, we propose PANG, a pattern-based generic framework that represents graphs as vectors, by identifying and leveraging their most discriminative subgraphs. We show how PANG, coupled with a standard classifier such as SVM, can detect fraud in public procurement, by applying it to an existing database (FOPPA). Traditional fraud detection approaches typically use tabular data to compute red flags to estimate risk, and fail when these data are incomplete. PANG leverages relational information between economical agents, and our experiments confirm that the use of graphs makes it possible to overcome this issue. They also show that prediction performance can be improved by mining closed or induced patterns, which constitute a set of predictors less redundant than general patterns. Finally, in this context, a clear advantage of PANG relies on the explainability of these discriminative patterns, which can be interpreted and associated with human behaviors such as favoritism.

%%%%%%%%%%%%%%%%%%%%%%%%%%%%%%%%%%%%%%%%%%%%%%%%%%%%%%%%%%%%%%%%%%
\bibliographystyle{splncs04}
\bibliography{potin_biblio.bib}

%%%%%%%%%%%%%%%%%%%%%%%%%%%%%%%%%%%%%%%%%%%%%
\appendix

\section{Classifier Comparison}
\label{app:classifier}
Table~\ref{tab:ClassifierChoice} complements the results presented in Section~\ref{sec:Benchmark}, by showing the performance obtained by a selection of classifiers on the FOPPA dataset, using the PANG\_GenBin representation and all available patterns.

\begin{table}[h!]
    \begin{center}
        \tabcolsep = 2\tabcolsep
        \begin{tabular}{l r r}
            \hline
            \textbf{Classifier} & \textbf{\textit{Anomalous} Class} & \textbf{\textit{Normal} Class} \\
            % & \textbf{$F$-Score} & \textbf{$F$-Score} \\
            \hline
            C-SVM & \textbf{0.93 (0.02)} & \textbf{0.93 (0.02)} \\
            Random Forest & 0.91 (0.02) & 0.91 (0.02) \\
            K Neighbors & 0.74 (0.04) & 0.71 (0.06)\\
            Gradient Boosting & 0.87 (0.02) & 0.87 (0.04) \\
            \hline
        \end{tabular}
        \caption{$F$-Scores for the \textit{Anomalous} and \textit{Normal} classes, according to all the considered classifiers.} 
        \label{tab:ClassifierChoice}
    \end{center}
\end{table}

%%%%%%%%%%%%%%%%%%%%%%%%%%%%%%%%%%%
\section{Ethical Implications}
Anomaly detection can have ethical implications, for instance if the methods are used to discriminate against certain individuals. In this respect, however, our PANG methodological framework does not present any more risk than the supervised classification methods developed in machine learning.

Moreover, this work takes place in the framework of a project aiming, among other things, at proposing ways of automatically red flagging contracts and economic agents depending on fraud risk. Therefore, the method that we propose is meant to be used by public authorities to better regulate public procurement and the management of the related open data.

Finally, the data used in this article are publicly shared, and were collected from a public open data repository handled by the European Union. They do not contain any personal information, and cannot be used directly to infer any personal information, as they only describe the economic transactions of companies and public institutions regarding public procurement.

\end{document}